\documentclass[10pt,twocolumn,letterpaper]{article}
\usepackage[pagenumbers]{cvpr}

\usepackage[pagebackref,breaklinks,colorlinks]{hyperref}

\usepackage[table,dvipsnames]{xcolor}
\usepackage{booktabs}
\usepackage{multirow}
\usepackage{makecell}
\usepackage{colortbl} 

\usepackage{graphicx}
\usepackage{subcaption}
\usepackage{tikz}
\usepackage{adjustbox}

\usepackage{amsmath} 
\usepackage{amsfonts} 
\usepackage{amssymb} 

\usepackage{pifont}
\usepackage{soul} 
\usepackage{changepage,threeparttable} 
\usepackage{tabularx}
\usepackage{paralist}
\usepackage{enumitem}
\usepackage{hhline}
\usepackage{wrapfig}
\usepackage{algorithm}
\usepackage{algpseudocode}

\definecolor{myred}{RGB}{174, 35, 23}

\newcommand{\myred}[1]{\textcolor{myred}{#1}}

\newcommand{\hgreen}[1]{\textcolor{ForestGreen}{#1}} 
\newcommand{\LeftComment}[1]{%
    \Statex \hspace{-\algorithmicindent} \(\triangleright\) \textit{#1}%
}
\definecolor{summary_color}{HTML}{C7E1F6} 
\definecolor{perception_color}{HTML}{B7D6C7}    
\definecolor{reasoning_color}{HTML}{D4D4FF}     
\definecolor{navigation_color}{HTML}{FFDE9A}     
\DeclareMathOperator*{\argmax}{arg\,max}

\title{\raisebox{-0.01\textheight}{\includegraphics[width=0.05\textwidth]{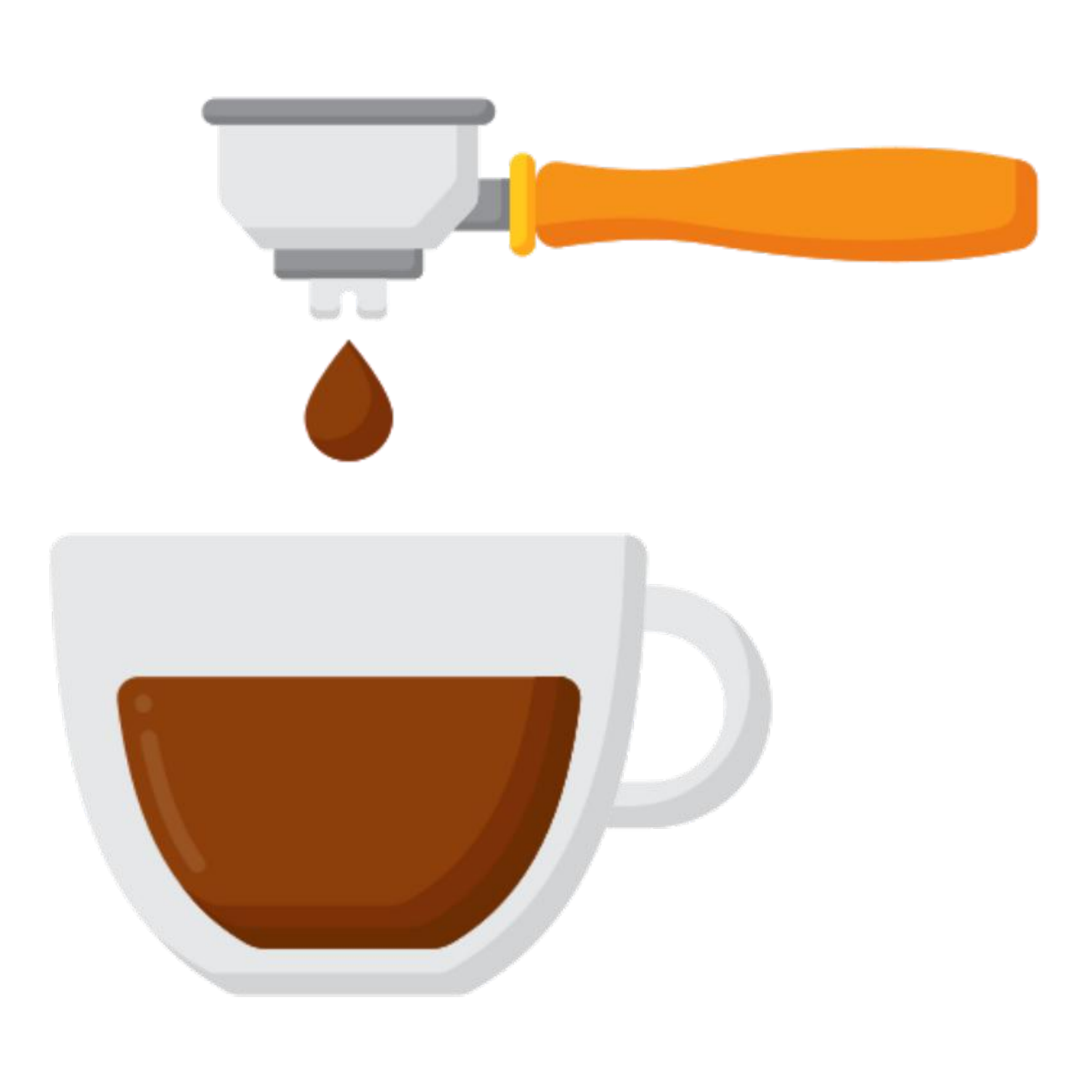}} VideoEspresso: A Large-Scale Chain-of-Thought Dataset for Fine-Grained Video Reasoning via Core Frame Selection}

\author{
    Songhao Han\textsuperscript{\rm 1},
    Wei Huang\textsuperscript{\rm 2},
    Hairong Shi\textsuperscript{\rm 1},
    Le Zhuo\textsuperscript{\rm 3},
    Xiu Su\textsuperscript{\rm 4},
    Shifeng Zhang\textsuperscript{\rm 5},
    Xu Zhou\textsuperscript{\rm 5},\\
    Xiaojuan Qi\textsuperscript{\rm 2}, 
    Yue Liao\textsuperscript{\rm 6}\footnotemark[2],
    Si Liu\textsuperscript{\rm 1}\footnotemark[2]
    \and  
  \textsuperscript{\rm 1}Beihang University
  \textsuperscript{\rm 2}The University of Hong Kong 
  \textsuperscript{\rm 3}Shanghai AI Lab
  \and 
  \textsuperscript{\rm 4}Central South University
  \textsuperscript{\rm 5}Sangfor Technologies Inc.
  \textsuperscript{\rm 6}CUHK
  \and
 \tt\small \{hshjerry,liusi\}@buaa.edu.cn, \tt\small\{aaron.weihuang, liaoyue.ai\}@gmail.com\\
    }

\begin{document}

\twocolumn[{
\maketitle
\vspace{-30pt}
\begin{center}
    \includegraphics[width=1\linewidth]{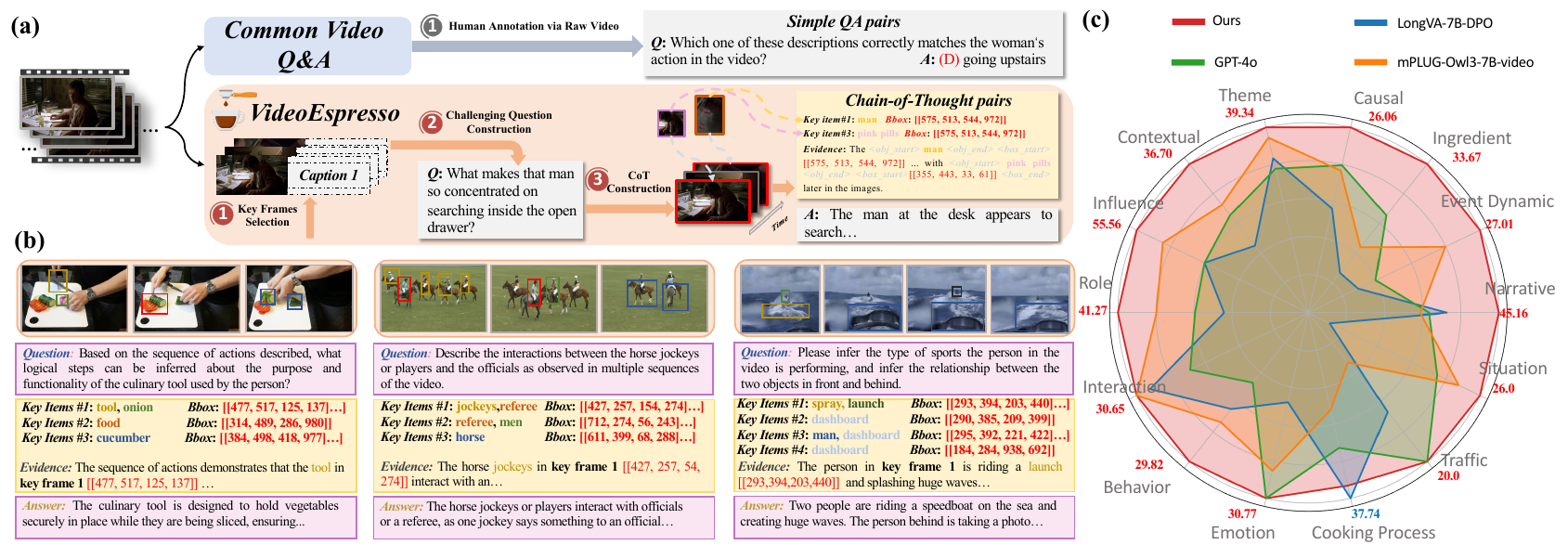}
    
    \vspace{-3mm}\captionof{figure}{\textbf{Overview of \textit{VideoEspresso}}. (\textbf{a)} Comparison of annotation pipelines: Unlike traditional videoQA datasets, \textit{VideoEspresso} features an automatic pipeline for constructing complex reasoning QA tasks and multimodal Chain-of-Thought (CoT) annotations. This enhances the diversity of QA data and significantly improves scalability. \textbf{(b)} Examples from \textit{VideoEspresso}: Illustrated are sample question-answer pairs, along with CoT bounding boxes and evidence annotations, demonstrating the dataset's richness. \textbf{(c)} Benchmark performance: Comparative results on our benchmark highlight the video reasoning capabilities of our model.}
\label{fig:fig1}
\end{center}
    }]

\renewcommand{\thefootnote}{\fnsymbol{footnote}}
\footnotetext[2]{Corresponding Author}
\begin{abstract}
The advancement of Large Vision Language Models (LVLMs) has significantly improved multimodal understanding, yet challenges remain in video reasoning tasks due to the scarcity of high-quality, large-scale datasets. Existing video question-answering (VideoQA) datasets often rely on costly manual annotations with insufficient granularity or automatic construction methods with redundant frame-by-frame analysis, limiting their scalability and effectiveness for complex reasoning. To address these challenges, we introduce \textit{VideoEspresso}, a novel dataset that features VideoQA pairs preserving essential spatial details and temporal coherence, along with multimodal annotations of intermediate reasoning steps. Our construction pipeline employs a semantic-aware method to reduce redundancy, followed by generating QA pairs using GPT-4o. We further develop video Chain-of-Thought (CoT) annotations to enrich reasoning processes, guiding GPT-4o in extracting logical relationships from QA pairs and video content. To exploit the potential of high-quality VideoQA pairs, we propose a Hybrid LVLMs Collaboration framework, featuring a Frame Selector and a two-stage instruction fine-tuned reasoning LVLM. This framework adaptively selects core frames and performs CoT reasoning using multimodal evidence. Evaluated on our proposed benchmark with 14 tasks against 9 popular LVLMs, our method outperforms existing baselines on most tasks, demonstrating superior video reasoning capabilities. Our code and dataset will be released at: \url{https://github.com/hshjerry/VideoEspresso}
\end{abstract}    
\vspace{-3mm}
\section{Introduction}
\label{sec:intro}

In recent years, the development of Large Vision Language Models (LVLMs)~\cite{Qwen-VL, liu2023llava,li2023blip, gao2023llamaadapterv2} has brought significant advancements to multi-modal understanding tasks. By integrating visual and language information through extensive data training, more advanced LVLMs families~\cite{openai2024gpt4o,li2024llava-ov} are able to generate reasonable outputs while fully leveraging the rich knowledge of LLMs, thus demonstrating excellent performance in tasks like image captioning and visual question answering. Recent research has started to explore extending LVLMs to the domain of video content understanding~\cite{li2023videochat,Maaz2023VideoChatGPT, li2023llamavid}. Although these efforts have shown great potential on certain basic video understanding benchmarks~\cite{li2023mvbench,yu2019activitynet,xiao2023nextgqa,mangalam2024egoschema}, their performance on complex video reasoning tasks remains less than satisfactory.

A primary limitation in video question-answering (VideoQA) research is the scarcity of high-quality, large-scale datasets. Current VideoQA datasets~\cite{xiao2021next,lei2018tvqa,yu2019activitynet} rely on costly manual annotations that often lack the granularity needed for detailed understanding, limiting scalability. Yet, LVLMs require vast amounts of multimodal QA pairs for effective training. Recently, advancements in large language models (LLMs) like GPT-4~\cite{openai2023gpt4} and Gemini-Pro~\cite{team2023gemini} have allowed for the automatic generation of QA pairs through carefully designed prompts. A straightforward approach is to use video metadata—typically high-level descriptions—and leverage LLMs to generate QA pairs based on this coarse information. However, the missing of crucial video details restricts the QA pairs' effectiveness for fine-grained reasoning. Alternatively, analyzing video frames for a more granular understanding is feasible, but video content is often redundant, with key information dispersed sparsely, making frame-by-frame analysis computationally expensive and prone to information overload.

To address these challenges, we propose a novel automatic VideoQA construction method and introduce a new dataset, \textit{VideoEspresso}. By preserving important spatial details with temporal coherence, we create a fine-grained reasoning-enabled VideoQA dataset that fosters more effective multimodal understanding. As shown in Fig.~\ref{fig:fig1}, we first design a semantic-aware key information extraction method to extract key information from the video. Unlike traditional methods that extract key frames based on image representations, we first map video frames to the linguistic space using an LVLM. We then remove similar frames based on semantic similarity, which reduces redundancy in the video data. To retain frame-level details and inter-frame correlation, we sequentially group the video frames and input them into GPT-4o~\cite{openai2024gpt4o}. With carefully designed prompts, we instruct the model to generate initial QA pairs and filter out low-quality data. To further expand the intermediate reasoning steps, we introduce video Chain-of-Thought annotations. We design prompts to guide GPT-4o in extracting logical relationship evidence from QA pairs and videos that are helpful for answers, including interactions of key objects in spatial and temporal flow. By annotating these logical processes, we ultimately aim to expand the chain of inference evidence on QA datasets.

To fully leverage the potential of the high-quality VideoQA pairs in our proposed \textit{VideoEspresso}, we introduce a novel framework, Hybrid LVLMs Collaboration for VideoQA, achieving cost-effective and accurate video LVLM reasoning. The framework is composed of a tiny \textit{Frame Selector} and a fine-grained reasoning LVLM. The \textit{Frame Selector} adaptively selects the most relevant core frames for the question based on the image-to-language mapping. These core frames are then submitted to the reasoning LVLM, where the model first extracts multimodal evidence based on the frame information, and ultimately provides the answer to the question through chain-of-thought reasoning, leveraging this evidence.
This dataset provides explicit annotations of key reasoning steps and image regions through both text and bounding boxes, which enables models to effectively use text and image localization information when answering questions.

Based on our dataset, we have constructed an evaluation benchmark that includes a set of GPT-4o-based open-ended evaluation metrics. We evaluated $9$ popular LVLMs as our comparison baselines. To assess video reasoning capabilities from different perspectives, we categorized the evaluation into $14$ tasks. Our method demonstrates significant advantages over baseline methods across most tasks.

\section{Related Work}
\label{sec:formatting}

\begin{figure*}[t]
   \centering
   \includegraphics[width=0.99\textwidth]{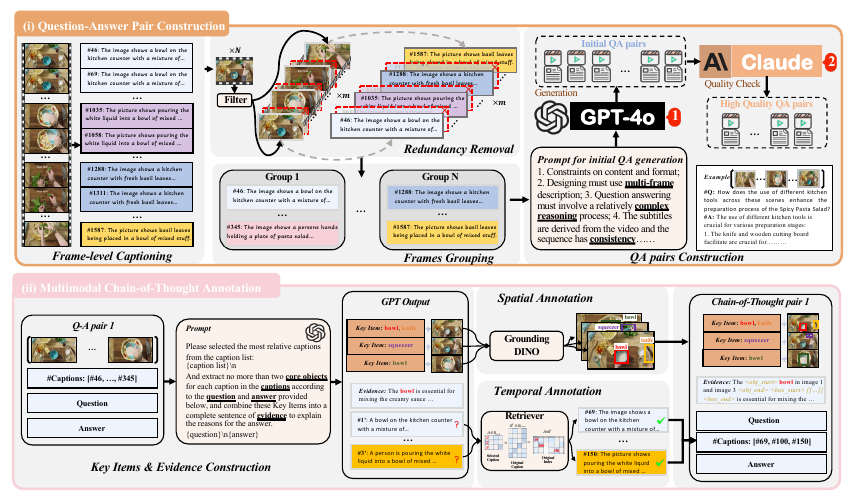} 
   \vspace{-4mm}
   \caption{\textbf{The automatic generation pipeline of \textit{VideoEspresso}.} (i) \textbf{Question-Answer Pair Construction}: We use video frame-leveled captions to extract the key frames of the video and group descriptions of these frames. Then, we prompt GPT-4 to design questions for each group of video frames. (ii) \textbf{Multimodal Chain-of-Thought Annotation}: We extract key evidence text and generate captions with the highest relevance to the question with GPT-4o. Additionally, we annotate spatial and temporal information for key items, which results in multimodal Chain of Thought data pairs grounded in both temporal and spatial dimensions.
       }\vspace{-3mm}
   \label{data_generation}
\end{figure*}

\noindent\textbf{VideoQA Dataset.} 
Traditional videoQA datasets~\cite{lei2018tvqa, xiao2023nextgqa, yu2019activitynet} rely heavily on manual annotations, where annotators watch videos, summarize content, and generate QA pairs based on set guidelines. This labor-intensive process limits scalability. With advancements in LLM capabilities~\cite{chatgpt, team2023gemini, claude3}, recent approaches use tailored prompts to utilize LLMs for annotation, often relying on metadata or detailed captions~\cite{chen2024panda, song_summarizing_web_videos} to construct QA data~\cite{li2023mvbench, song2023moviechat, timechat}. However, these methods often lack fine-grained video information and depend heavily on raw video data. In contrast, we introduce an automatic pipeline for QA pair generation that processes and annotates raw data without manual input, enhancing scalability.

\noindent\textbf{Video LVLMs.} 
In recent years, large vision-language models (LVLMs) have advanced VideoQA tasks significantly. Prior works have enhanced model performance by experimenting with various architectures, with some aligning visual and textual features through Q-Former~\cite{damonlpsg2023videollama,li2024llava-ov,chen2024sharegpt4video}, while others concatenate frame-level features directly~\cite{li2023videochat,jin2023chatunivi}. Balancing frame count with token efficiency has become a key focus, typically addressed by uniform sampling~\cite{zhang2024longva,chen2024sharegpt4video} or additional modules~\cite{song2023moviechat, li2023llamavid}. Our approach employs a tiny model that selectively captures frames relevant to the question, minimizing context length while preserving essential spatiotemporal information.

\noindent\textbf{Visual CoT.}
Chain-of-Thought~(CoT) techniques enhance the reasoning abilities of LLMs by guiding them through intermediate reasoning steps to produce more accurate answers. While prior works~\cite{zhang2023multimodalcot, yang2023mmreact} have applied CoT to visual tasks, they primarily focus on text-level reasoning, often overlooking visual comprehension. Recent studies~\cite{wu2024vstar, shao2024visualcot} improved performance by targeting specific image regions. In videoQA, VideoCoT~\cite{wang2024videocot} focuses on text-level reasoning, while VoT~\cite{fei2024videocoticml} emphasizes spatial relationships. Our approach integrates key object regions and core frames in the video CoT reasoning, capturing both spatial and temporal details to enhance video understanding.
\section{VideoEspresso}
\label{sec:dataset}
In this section, we introduce \textit{VideoEspresso}, a large-scale VideoQA dataset designed to facilitate high-level reasoning over macroscopic video semantics. This dataset is generated through a scalable, fully automated generation pipeline that produces high-quality reasoning VideoQA pairs from distilled video content. The \textit{VideoEspresso} construction pipeline consists of the following key stages: (1) We collect raw video data, followed by redundancy reduction streamline video frames that encapsulate essential content; (2) Based on these frames, we generate QA pairs that capture the core semantics of each video; (3) To further enable interpretability and bolster the benchmark for complex reasoning capabilities, we incorporate fine-grained Chain-of-Thought (CoT) annotations, which connect core visual elements through spatial and temporal interactions, bridging the reasoning gaps in traditional VideoQA pairs. 

\subsection{Video Data Curation}
\label{subsec:data_curation}
\begin{figure}[t]
   \centering
   \includegraphics[width=0.45\textwidth]{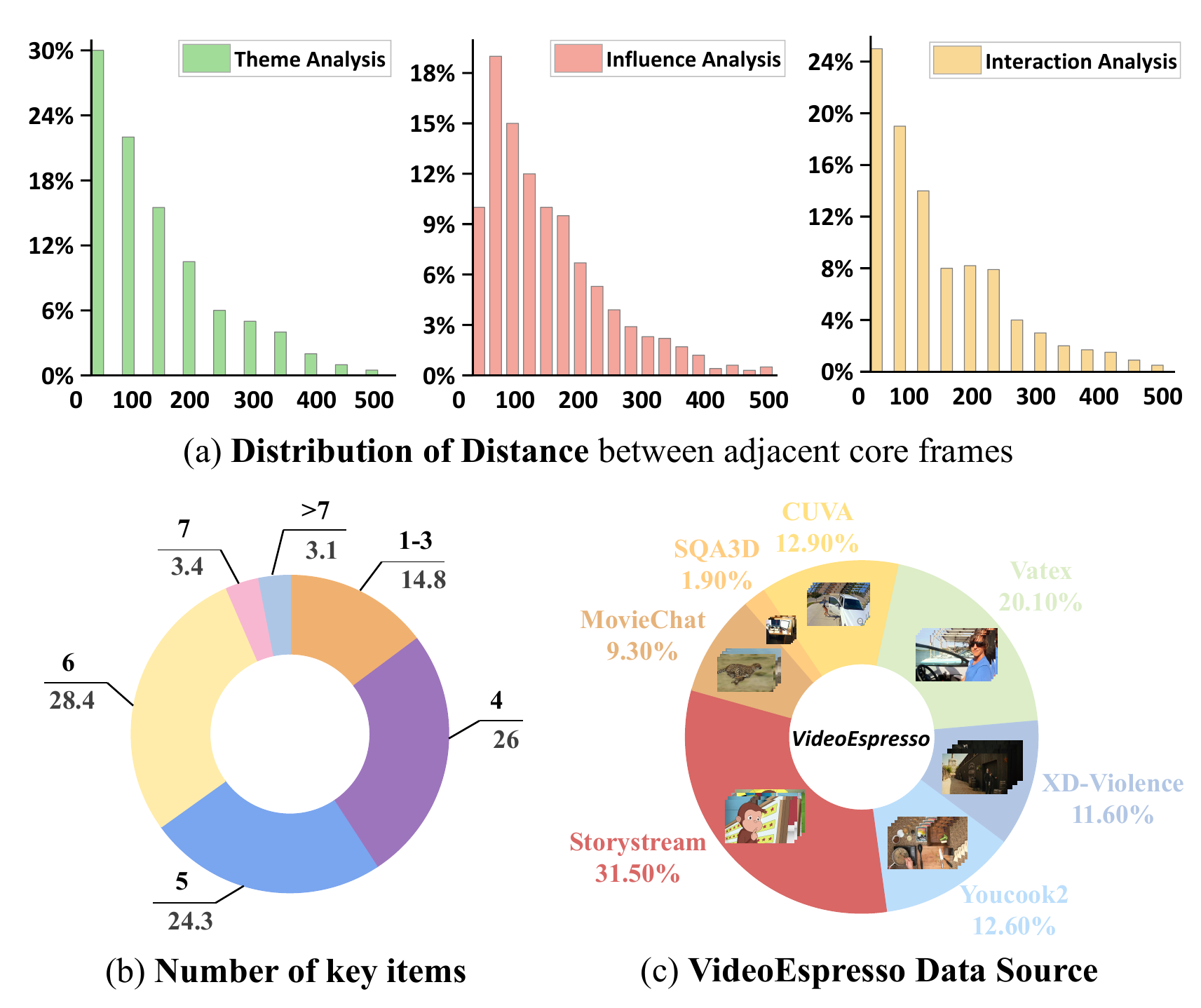} 
   \vspace{-3mm}
   \caption{\textbf{The statistical analysis of our \textit{VideoEspresso} dataset.}
   }\vspace{-3mm}
   \label{fig:data_fig1}
\end{figure}

We leverage the vast amount of unannotated Internet videos for building scalable datasets.
To build a videoQA dataset with complex semantic reasoning, selecting appropriate data sources and types is essential. As illustrated in Fig.~\ref{fig:data_fig1}, we collect raw videos from 7 datasets~\cite{song2023moviechat,zhou2018towards,wang2019vatex,du2024uncovering,yang2024seedstory,ma2022sqa3d,wu2020not} encompassing rich temporal dynamics, logical sequences, and causal relationships. These high-level semantics provide a strong foundation for constructing a complex and coherent question-answering dataset. We conduct a manual review of these videos to evaluate their reasoning potential. The dataset includes diverse video types, covering genres such as news, movies, documentaries, animation, and educational content. Based on these characteristics, we predefined $14$ tasks to assess the model's capabilities across various dimensions of video reasoning tasks.

\subsection{Redundancy Removal in Video Frames}
\label{subsec:frames selection}

The goal of this module is to eliminate redundant information in the video and retain the essential content by selecting a concise sequence of frames. Different videos exhibit varying rates of content and scene changes. Initially, we determine an appropriate sampling interval based on the type of video. For instance, for rapidly changing dynamic scenes, we set the FPS between 2 and 4, whereas for static scenes, we choose a longer sampling interval, with FPS set to 1. Then, to capture the fine-grained semantic information of the video as input for constructing QA pairs, we use InternVL2-8B~\cite{chen2023internvl} to perform frame-level captioning on all sampled frames. To filter out redundant frames in the video, we leverage the language retrieval model BGE-M3~\cite{chen2024bge} to preliminarily remove highly similar frames through fine-grained semantic filtering. Specifically, for all sampled frame descriptions $\mathcal{C}$, if the cosine similarity between the textual features $\phi_T(c)$ of adjacent captions exceeds a preset threshold $\tau$, we apply a Last-In-First-Out (LIFO) filtering approach. This process results in a concise caption sequence $\mathcal{C}'$ and the corresponding frames.
The process is formulated as follows:
\begin{align}
    \mathcal{S} &= \argmax_{c_i,c_j\in \mathcal{C}} \cos(\phi_T(c_i), \phi_T(c_j)),\\
    \mathcal{C}&\longrightarrow\mathcal{C}'~(c\in \mathcal{C}', \text{if}~\mathcal{S}(c)<\tau),
\end{align}
where $\mathcal{S}$ denotes the similarity matrix, and $\phi_T(c)$ represents the feature computation for the caption $c$.

\subsection{Question-Answer Pair Construction}
\label{subsec:qa construction}

This module aims to leverage the powerful language reasoning capabilities of LLMs to automatically construct high-quality video reasoning QA pairs based on detailed descriptions of video frames. To maintain semantic continuity within the groups and avoid issues such as model hallucinations and failure to follow instructions—caused by an excessive number of tokens—we adopt a continuous grouping approach to streamline frames. Specifically, for all captions $\mathcal{C}'$ of a single video, every 15 consecutive frame captions are grouped into a group $\mathcal{G}_i$ that preserves both frame-level details and inter-frame correlations.

\begin{figure}[t]
   \centering
   \includegraphics[width=0.475\textwidth]{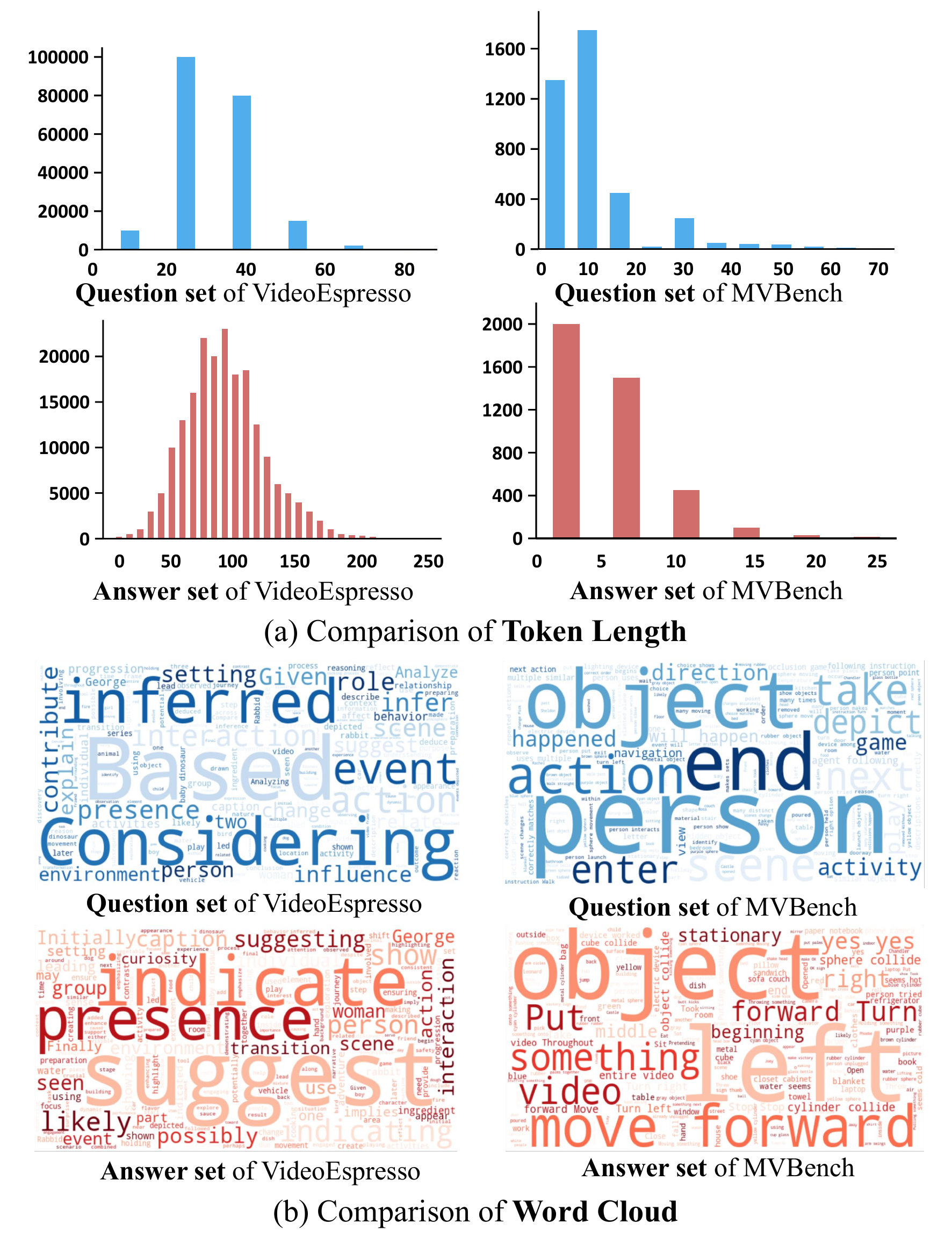} 
   \vspace{-6mm}
   \caption{\textbf{The dataset attributes comparison between our \textit{VideoEspresso} and MVbench.}
   }\vspace{-5mm}
   \label{fig:data_fig2}
\end{figure}

To obtain complex reasoning QA pairs, we design and iteratively refine prompts to ensure that the LLM follows the rules in constructing them. As shown in the top-right corner of Fig.~\ref{data_generation}, our prompts instruct GPT-4o~\cite{openai2024gpt4o} to generate question-answer pairs based on multi-frame descriptions, while also ensuring that GPT-4o maintains consistency between the descriptions to construct complex video reasoning questions. 
To improve the quality of the QA pairs, we validate them by designing an additional LLM~\cite{claude3} to verify the quality of both the questions and answers, including eliminating hallucinations in the QA pairs, checking the factual accuracy of the answers, and filtering out answers to highly subjective or difficult-to-evaluate open-ended questions. Ultimately, we obtain high-quality reasoning QA pairs for these videos and record the frame sequence grouping corresponding to each QA pair.

\begin{figure*}[t]
   \centering
   \includegraphics[width=\textwidth]{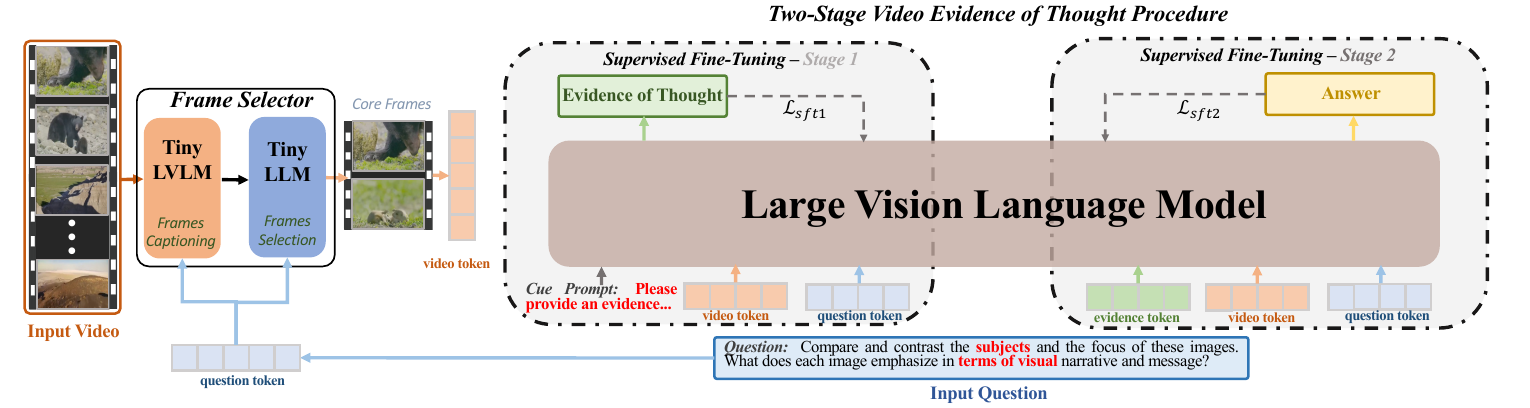} 
   \vspace{-2mm}
   \caption{\textbf{Two-Stage Video Evidence of Thought Training Procedure.} The Frame Selector comprises a tiny LVLM and a tiny LLM, tasked with generating captions for videos and selecting the most relevant frame to as core video token for large reasoning model. A two-stage supervised fine-tuning technique is employed. During stage-1, a set of cue prompts is introduced to guide the model in producing evidence, while in stage-2, the evidence generated from stage-1 is concatenated and used directly to guide the answer generation.}
   \label{pipeline}
\vspace{-0.1in}
\end{figure*}

\subsection{Multimodal Chain-of-Thought Annotation}
\label{subsec:cot}

To further enhance the reasoning capabilities of models, this module focuses on annotating multimodal evidence that contains key spatiotemporal information. First, we group the Q-A pairs obtained in Sec.~\ref{subsec:qa construction} along with the corresponding frame sequences as input, and design the prompt shown in the lower-left corner of Fig.~\ref{data_generation} to guide GPT-4o~\cite{openai2024gpt4o} in extracting key information. Sparse core frames are sufficient to capture enough information to answer the question; hence, we need to obtain the core frames most relevant to the question. Accordingly, our designed prompt primarily guides GPT-4o to extract the following key information: (1) From the captions group, select the captions most relevant to the question, i.e., the captions of core frames; (2) extract key objects from these captions, i.e., the key items; and (3) organize these key objects into a natural language description as evidence to support answering the question, i.e., the evidence. To expand the dimensions of reasoning, we annotate these key elements with temporal and spatial information. For spatial annotation, we apply GroundingDINO~\cite{Liu2023GroundingDM} to mark bounding boxes around all key items and leverage CLIP-ViT-B/32~\cite{radford2021clip} to verify consistency between labels and objects within the bounding boxes. For temporal annotation, since GPT-4o's generated captions of core frames $\mathcal{G}_{GPT}$ in (1) cannot directly match the original captions at the string level, we employ BGE-M3~\cite{chen2024bge} to retrieve the caption in the original set \(\mathcal{G}_i\) and obtain temporal grounding information \(t\).
The process is formulated as follows:
\begin{align}
t = \arg\max_{k}\cos(\phi_T(c_j), \phi_T(c_k)),
\end{align}
where $c_j \in \mathcal{G}_{GPT}$, $c_k \in \mathcal{G}_i$.

Finally, we obtain both textual evidence and multimodal key information that includes temporal and spatial dimensions, which can serve as the intermediate reasoning step to assist in answering questions.

\subsection{Data Analysis}
\label{subsec:data_analysis}

In Fig.~\ref{fig:data_fig1} and Fig.~\ref{fig:data_fig2}, we provide a visualization of the data analysis and comparisons. To investigate the temporal distribution of key information in videos, we first examine the distribution of distances between adjacent key frames across different tasks. As shown in Fig.\ref{fig:data_fig1}.(a), the distance distribution between key frames varies significantly across tasks, indicating that the conventional strategy of uniformly sampling video frames is suboptimal and introduces substantial redundancy. As shown in Fig.~\ref{fig:data_fig1}.(b), the number of key items in the CoT of our dataset is diverse, encompassing a range from a few to numerous critical elements, reflecting the complexity and diversity of the visual content.
In addition to the unique characteristics of our dataset, we also compare it with the QA content of a popular dataset, MVBench~\cite{li2023mvbench}. 
As shown in Fig.~\ref{fig:data_fig2}.(a), we illustrate the differences between the datasets in terms of token length. The QA set length in MVBench (right) is shorter, while the answer set in our \textit{VideoEspresso} (left) is on average much longer and exhibits greater diversity in distribution.
As illustrated in Fig.~\ref{fig:data_fig2}.(b), we further present a comparison of the word clouds for \textit{VideoEspresso} and MVBench. In the Question set, our \textit{VideoEspresso} emphasizes reasoning based on visual facts, with keywords like ``\textit{considering}", ``\textit{based}" and ``\textit{inferred}" In contrast, MVBench emphasizes basic inquiries such as ``\textit{object}", ``\textit{person}" and ``\textit{action}" In the Answer set, \textit{VideoEspresso} includes not only reasoning-related keywords as previously described but also terms associated with reasoning steps, such as ``\textit{Initially}" and ``\textit{Finally}" On the other hand, MVBench focuses on object definitions and spatial relationships within videos, with keywords like ``\textit{object}", ``\textit{left}" and ``\textit{forward}".

\section{Hybrid LVLMs Collaboration for VideoQA}

\label{sec: method}
To fully unleash the potential of high-quality video QA pairs offered by \textit{VideoEspresso}, we propose an efficient video reasoning framework with hybrid LVLMs collaboration to enable cost-effective and accurate video LVLM reasoning. As shown in Fig.~\ref{pipeline} the framework consists of two core components: a lightweight selector that identifies core frames that are closely related to the input question, and a powerful LVLM that performs content understanding and reasoning based on these selected core frames.

\subsection{Core Frames Selection via Tiny LVLM}
We propose a Lightweight Selector designed to extract core frames that are closely aligned with the question from input videos. Unlike traditional keyframe extraction methods, which primarily filter out semantically similar frames, our approach dynamically selects question-driven core frames to meet diverse task requirements. This enables a reduction in the number of frames passed to large models compared to conventional methods that rely solely on frame semantic similarity. Moreover, this selector serves as a plug-and-play module that can be inserted before any LVLMs.

Our architecture comprises a lightweight LVLM with 1 billion parameters and an LLM with 0.5 billion parameters in a sequential setup. The LVLM's function is to convert video frames into language descriptions, while the LLM selects frames most relevant to the question based on these descriptions. Specifically, the process consists of two steps. 

(1) Frame Captioning: Given a video $V$ and a specified frames-per-second (\texttt{FPS}) sampling rate, the \texttt{LVLM} samples frames and generates a caption $c_i$ for each frame $f_i$. This process can be formulated as:
\begin{align}
   \{f_i\}_{i=1}^{N} &= \texttt{SampleFrames}(V, \texttt{FPS})\\
   \{c_i\}_{i=1}^{N} &= \texttt{LVLM}(\{f_i\}_{i=1}^{N})
\end{align}
where $N$ is the total number of sampled frames and $\{c_i\}_{i=1}^{N}$ represents the collection of captions for these frames.

(2) Core Frame Selection: Using the set of captions $\{c_i\}_{i=1}^{N}$ and a question $q$, the \texttt{LLM} identifies the subset of captions most relevant to $q$, resulting in a set of core frame captions $\{c_j'\}_{j=1}^{M}$:
\begin{align}
   \{c_j'\}_{j=1}^{M} = \texttt{LLM}(\{c_i\}_{i=1}^{N}, q)
\end{align}
where $M \leq N$ and $\{c_j'\}_{j=1}^{M}$represents the final set of core captions selected for their relevance to the question. 
Since this step requires minimal reasoning from the model, we adopt a cost-efficient solution to address the challenges posed by excessive token length when handling video input in large models.

\subsection{Fine-Grained Reasoning via LVLM}
Given key frames extracted from the first stage, our goal is to enable the model to effectively leverage multimodal spatiotemporal evidence for answering complex reasoning tasks. We design a two-stage supervised fine-tuning paradigm. In the first stage, we guide the model to extract essential visual evidence from video data relevant to the question, establishing a foundation for deeper reasoning. This is achieved through supervised fine-tuning using instructions like \textit{``Please provide evidence to help answer the question."} to guide the model in generating evidence. 
This evidence-based generation process not only filters core information but also enhances multimodal alignment and prepares the model for subsequent reasoning tasks.

In the second stage, we further fine-tune the model to directly generate answers based on the extracted multimodal evidence.
This is achieved through supervised fine-tuning using instructions like \textit{``Please answer the question with the help of evidence."} to guide the model in answering with evidence. Unlike traditional single-stage question-answering methods, this two-stage structure divides evidence generation and answer generation, enhancing transparency in reasoning and boosting response accuracy. Besides, it ensures that the model gradually integrates multimodal information for complex spatiotemporal reasoning, significantly improving performance in video question-answering tasks by producing more logically coherent answers.

\subsection{Inference}
During the inference phase, we first use the Lightweight Selector to extract core frames from the video that are closely related to the question, serving as input for subsequent reasoning. We then leverage a fine-grained reasoning LVLM, which generates evidence through a chain-of-thought process to support the final answer generation. This workflow enables efficient question answering, from frame selection to answer generation.
\definecolor{front-color}{HTML}{F5FFFA}
\begin{table*}[tp]
    \centering
    \renewcommand{\arraystretch}{1.2}
    \setlength\tabcolsep{1pt}
    \resizebox{1.0\textwidth}{!}{
        \begin{tabular}{l|c|c|c|c|c|c|c|c|c|c|c|c|c|c|c|c|c|c}
        \Xhline{1.0pt}
       \textbf{Models} & \textbf{\#Frames} & \textbf{Param} & \textbf{TFLOPs} & \textbf{Narra.} & \textbf{Event} & \textbf{Ingre.} & \textbf{Causal} & \textbf{Theme} & \textbf{Conte.} & \textbf{Influ.} & \textbf{Role} & \textbf{Inter.} & \textbf{Behav.} & \textbf{Emoti.} & \textbf{Cook.} & \textbf{Traff.} & \textbf{Situa.} & \textbf{Avg.} \\
        \midrule
        \rowcolor{Gray}
        \multicolumn{19}{c}{\textit{Closed-source LVLMs}} \\
        \midrule
        GPT-4o~\cite{openai2024gpt4o} & FPS=3 & - & - &32.3  & 16.7  & 25.5  & 22.8  & 32.8  & 27.5  & 37.5  & 28.6  & 24.2  & 19.3  & 30.8  & 30.2  & 20.0  & 22.0  & 26.4  \\
        Qwen-VL-Max~\cite{Qwen-VL} & FPS=3 & - & - &33.9  & 22.4  & 23.5  & 21.4  & 26.2  & 30.3  & 41.7  & 30.2  & 27.4  & 26.3  & 20.0  & 20.8  & 16.7  & 24.0  & 26.0  \\
        \midrule
        \rowcolor{Gray}
        \multicolumn{19}{c}{\textit{Opened-source LVLMs}} \\
        \midrule
        LLaVA-1.5~\cite{liu2023llava15} & 4 & 7B & 14.50 & 32.3  & 21.3  & 19.4  & 17.1  & 26.2  & 20.2  & 36.1  & 33.3  & 21.0  & 21.1  & 20.0  & 35.8  & 16.7  & 18.0  & 24.2  \\
        InternVL2~\cite{chen2023internvl} & FPS=1 & 8B & 73.23 & 33.9  & 24.1  & 27.6  & 24.4  & \textbf{42.6}  & 33.0  & 45.8  & 28.6  & 19.4  & 22.8  & 21.5  & 34.0  & 20.0  & 24.0  & 28.7  \\
        LLaVA-N-Inter~\cite{li2024llava} & FPS=1 & 7B & 62.78 &24.2  & 23.6  & 26.5  & 19.2  & 31.1  & 32.1  & 31.9  & 17.5  & 24.2  & 21.1  & 26.2  & 30.2  & 13.3  & 20.0  & 24.4  \\
        Qwen2-VL~\cite{Qwen-VL} & FPS=1 & 7B & 64.60 &27.4 & 23.0 & 24.5 & 23.5 & 29.5 & 31.2 & 47.2 & 31.7 & 22.6 & 28.1 & 40.0 & 22.6 & 30.0 & 18.0 & 28.5 \\
        LongVA-DPO~\cite{zhang2024longva} & 128 & 7B & 465.4 &35.5  & 14.9  & 16.3  & 19.0  & 34.4  & 22.0  & 37.5  & 23.8  & 29.0  & 22.8  & 20.0  & \textbf{37.7}  & 16.7  & 12.0  & 24.4  \\
        mPLUG-Owl3~\cite{ye2024mplug} & FPS=1 & 7B & 89.78   &30.6  & 23.6  & 20.4  & 22.3  & 37.7  & 29.4  & 48.6  & 34.9  & 30.6  & 24.6  & 27.7  & 24.5  & 13.3  & 24.0  & 28.0  \\
        LLaVA-N-Video~\cite{zhang2024llavanextvideo} & FPS=1 &7B & 60.42 & 31.2 & 20.2 & 16.2 & 17.6 & 36.5 & 32.7 & 30.6 & 24.5 & 26.4 & 24.5 & 34.7 & 20.8 & 20.3 & 17.0 & 25.2 \\
        \hline
       Ours & 2.36 & 8.5B & \textbf{9.26} &\textbf{45.2} & \textbf{27.0} & \textbf{33.7} & \textbf{26.1} & 39.3 & \textbf{36.7} & \textbf{55.6} & \textbf{41.3} & \textbf{30.6} & \textbf{29.8} & \textbf{30.8} & 35.8 & \textbf{20.0} & \textbf{26.0} & \textbf{34.1}  \\
       \bottomrule
        \end{tabular}
    }
    \vspace{-0.2cm}
    \caption{
    \textbf{Main Result on Our Objective Benchmark.} We report results of closed-source and opened-source LVLMs with ours. The process of constructing task evaluations is shown in the supplementary. TFLOPs refers to the total computational cost of inference, measured under the same 16-second video input.
   }
    \label{tab:bench}
    \vspace{-0.3cm}
\end{table*}
\section{Experiments}
\label{sec:exp}
\begin{table}[t]
    \small
    \begin{center}
    \vspace{5pt}
    \renewcommand{\arraystretch}{0.5}
    \scalebox{1.}{
    \begin{tabular}{{@{}l | c c c c |c @{}}}
    \toprule
Models & Log. & Fac.& Acc. & Con. & Overall\\
    \midrule
    \rowcolor{Gray}
    \multicolumn{6}{c}{\textit{Closed-source LVLMs}} \\
    \midrule
GPT-4o & 73.15 & 63.11 & 61.66 & 70.02 & 66.13\\
Qwen-VL-Max & 62.46 & 50.33 & 48.43 & 60.21 & 53.37\\
    \midrule
    \rowcolor{Gray}
    \multicolumn{6}{c}{\textit{Open-source LVLMs}} \\
    \midrule
LLaVA~1.5 & 60.53 & 49.56 & 49.93 & 62.1 & 52.12\\
InternVL2 & 70.64 & 56.32 & 54.53 & 66.76 & 60.05\\
LLaVA-N-inter & 63.27 & 52.34 & 48.45 & 66.78 & 55.16\\
Qwen2-VL-7B & 66.31 & 53.67 & 50.84 & 68.88 & 57.66\\
LongVA-7B-DPO & 67.98 & 54.72 & 52.78 & 58.38 & 57.19\\
mPLUG-Owl3 & 66.14 & 53.05 & 50.97 & 67.3 & 57.14\\
LLaVA-N-Video & 63.42 & 54.11 & 49.55 & 63.31 & 56.43\\
    \midrule
Ours & \textbf{72.25} & \textbf{61.28} & \textbf{59.68} & \textbf{75.73} & \textbf{65.84}\\
    \bottomrule     
    \end{tabular}}
    \vspace{-2mm}
    \caption{\textbf{Results on Subjective Benchmark.} We report the metrics of Logic (Log.), Factuality (Fac.), Description Accuracy (Acc.), and Conciseness (Con.).} 
    \vspace{-10pt}
    \label{tab:sub_bench}
    \end{center}
\end{table}

\subsection{Overview of the Evaluation Benchmark}
Our \textit{VideoEspresso} includes 14 predefined tasks, with each constructed QA pair matched to a corresponding task using GPT-4o.  If no suitable task alignment is identified, the pair is categorized as ``\textit{Others}". To establish a comprehensive benchmark, the defined tasks encompass diverse perspectives, including time, logic, scene, behavior, and state, illustrated by examples such as ``\textit{Event Dynamics}", ``\textit{Causal Analysis}" and ``\textit{Theme Analysis}". Additionally, the framework incorporates real-world application tasks, such as ``\textit{Cooking Process}" and ``\textit{Traffic Analysis}". The benchmark assesses the performance of LVLMs through both objective and subjective evaluations, providing a multifaceted analysis of their capabilities.

\noindent\textbf{Experimental Setting.}
To comprehensively evaluate the capabilities of LVLMs on VideoQA tasks, we selected: (1) closed-source large models, such as GPT-4o~\cite{openai2024gpt4o} and Qwen-VL-Max~\cite{Qwen-VL}; (2) general-purpose LVLMs that claim strong video capabilities on video benchmarks, such as InternVL~\cite{chen2023internvl} and Qwen2-VL~\cite{Qwen-VL}; and (3) popular video LVLMs, such as LongVA~\cite{zhang2024longva} and mPLUG-Owl3~\cite{ye2024mplug}.To ensure the fairness of the reported accuracies, the video frame sampling scheme, temperature, and other parameters follow the settings from the original paper. Additionally, we standardize the maximum token length of the outputs to $512$. As our model training details, the learning rate is set to $2$e-$5$, the warmup rate is $0.03$, and we train the model for one epoch with global batch size of $16$. The training and evaluation process is facilitated on 8 NVIDIA-A100 GPUs.

\noindent\textbf{Evaluation.} 
To more accurately evaluate the open-ended responses of LVLMs, we propose a two-step evaluation method based on fine-grained semantic similarity. In the first step, we assess the semantic similarity between the model's output and the reference answer. If the similarity exceeds $80\%$, the output is considered potentially correct from a semantic perspective. In the second step, we introduce three highly confounding distractors for each reference answer. We then calculate the similarity between the model's output and each distractor. If the similarity with any distractor surpasses that of the model's output to the reference answer, the response is deemed incorrect. Only if the output passes both steps is it classified as correct.

Meanwhile, we incorporate subjective evaluation by assessing the generated content across multiple dimensions, including logic, factuality, desciption accuracy, and conciseness. To facilitate this, we designed an evaluation framework that prompts GPT-4o~\cite{openai2024gpt4o} to rate the model's output on a scale of 1 to $10$ based on the ground truth, along with providing an overall score. Moreover, during evaluation, we do not directly input the options into the model, which effectively prevents potential information leakage. Finally, we report the accuracy for each task across the entire \textit{VideoEspresso} dataset.

\begin{table*}[t]
\centering
\begin{minipage}[t][3cm][t]{0.23\linewidth}
\small
\centering
\begin{adjustbox}{width=0.9\linewidth}
\begin{tabular}{@{}lc@{}}
\toprule
~\textbf{Setting} & \textbf{Acc.}\\
\midrule
~baseline & 34.13 \\
~GT-CoT & $72.95_{\hgreen{+38.82}}$ \\
~w/o Bbox & $33.14_{\myred{-0.99}}$ \\
~w/o CoT & $31.32_{\myred{-2.81}}$ \\
\bottomrule
\end{tabular}
\end{adjustbox}
\vspace{-1mm}
\caption{\textbf{Ablation Studies on different CoT strategies.} \textit{GT} is 'ground truth'.}
\vspace{-8pt}
\label{table:ablation}
\end{minipage}\hspace{-2mm}
\hfill
\begin{minipage}[t][3cm][t]{0.68\linewidth}
\small
\centering
\begin{adjustbox}{width=0.9\linewidth}
\begin{tabular}{@{}lccccc@{}}
\toprule
\textbf{Selector} & \textbf{\#Frame} & \textbf{Add.}  & \textbf{GPU hr.} & \textbf{Inference Memory} & \textbf{Acc.}\\ \midrule
Uniform & 8 & - & - & 0G + 14G + 40G & $33.74$ \\
GT & 2.98 & - & -  & 0G + 14G +15G & $37.54_{\hgreen{+3.80}}$ \\
1B/1.5B & 2.77 & 2.5B & 1.33 & \myred{5G} +14G +\hgreen{14G} & $34.76_{\hgreen{+1.02}}$ \\
1B/0.5B & 2.36 & 1.5B & 0.37 & \myred{3G} +14G +\hgreen{12G} & $34.13_{\hgreen{+0.29}}$ \\
\bottomrule
\end{tabular}
\end{adjustbox}
\vspace{-1mm}
\caption{\textbf{Ablation studies on Selector.} GT refers to the ground truth time of the core frames annotation. ``1B/1.5B" represents the selector consists of InternVL2-1B and QwenLM-1.5B, and similarly for ``1B/0.5B". Add. stands for additional parameters.}
\label{table:selector}
\end{minipage}
\vspace{26pt}
\end{table*}

\subsection{Results on Benchmark}
\noindent\textbf{Objective Evaluation Results.} 
We evaluated $7$ open-source and $2$ closed-source LVLMs on $14$ video reasoning tasks in objective evaluation. 
As shown in Tab.~\ref{tab:bench}, our method achieves the state-of-the-art performance across $12$ tasks, and an average accuracy of $34.1\%$. This performance surpasses that of the top-performing open-source model, InternVL2~\cite{chen2023internvl}, and the closed-source GPT-4o~\cite{openai2024gpt4o} by $5.4\%$ and $7.7\%$, respectively. Compared to our selected backbone, LLaVA-Next-interleave, the performance improves by nearly $10\%$ after fine-tuning with reasoning instructions. In addition to its advantage in video reasoning QA, our method also demonstrates the leading efficiency. Specifically, the average number of input frames is only $1.8\%$ of that used by LongVA-DPO~\cite{zhang2024longva}, and the FLOPs calculated on the same video input are only $14.74\%$ of those for LLaVA-Next-interleave. Notably, InternVL2 and LongVA-DPO excel in the “Theme Analysis” and “Cooking Process” tasks, which is likely due to their exposure to large size of the same type of data during training process.

\noindent\textbf{Subjective Evaluation Results.} 
We evaluated the quality of LVLMs' answers in subjective evaluation across four aspects, including logic consistency, factuality, accuracy, and conciseness. As shown in Tab.~\ref{tab:sub_bench}, the results align closely with the observations from the objective evaluation. GPT-4o exhibits strong performance in both logical reasoning and factual accuracy, due to its robust language reasoning capabilities and extensive prior knowledge. However, among all open-source LVLMs, our method outperforms the approaches presented in the table across all four dimensions. Notably, in the Conciseness evaluation, our method surpasses GPT-4o by $5\%$, further demonstrating the significant contribution of our \textit{VideoEspresso} dataset in enhancing models' ability to learn video reasoning.

\subsection{Ablation Study}
\noindent\textbf{Ablation on CoT.}
Our results in Tab.~\ref{table:ablation} further prove the effectiveness of visual grounding in CoT evidence. Moreover, ablation experiments involving the ground truth of CoT and the ablation of the CoT process further demonstrate the potential of CoT in enhancing performance on visual tasks. The performance boost achieved by the CoT ground truth is significant, highlighting the importance of endowing LVLMs with reasoning QA capabilities.

\noindent\textbf{Ablation on Selector.}
As shown in Tab.~\ref{table:selector}, we conducted ablation experiments under different selector combination settings, specifically InternVL2-1B~\cite{chen2023internvl} $+$ QwenLM-1.5B~\cite{bai2023qwen} and InternVL2-1B $+$ QwenLM-0.5B. In this set of experiments, our core frame selection significantly improved video understanding capabilities compared to the uniform sampling method. Although selectors of different sizes may increase memory usage by 3GB or 5GB, the optimization of redundant keyframe tokens led to a reduction of 26-28GB in memory usage, resulting in a significant improvement in overall video understanding efficiency. We also tested more lightweight LVLMs, such as LLaVA-Next-interleave-0.5B~\cite{li2024llava} and Qwen-VL-1.5B~\cite{Qwen-VL}. However, the results did not meet expectations, likely due to the Tiny LVLMs processing too many image tokens, exceeding their capacity for handling them effectively.
\subsection{Adapting Selector to other LVLMs}
\definecolor{front-color}{HTML}{F5FFFA}

\begin{table}[t]
\small
    \centering
    \setlength\tabcolsep{1pt}
        \begin{tabular}{l|c|c|c|c|c}
        \Xhline{1.0pt}
        \textbf{Model} & \textbf{Sample} & \textbf{\#Frame} & $\textbf{Ratio}_{tok}$ & \textbf{TFLOPs} & \textbf{Acc.} \\
        \Xhline{1.0pt}
        GPT-4o & Uniform & 16 & 1 & - & 26.86 \\
        GPT-4o & 1B/0.5B & 2.77 & 0.17 & - & \hgreen{28.26} \\
        GPT-4o & 1B/1.5B & 2.36 & 0.15 & - & \hgreen{29.45} \\
        \midrule
        InternVL2 & Uniform & 16 & 1 & 73.23  & 28.57 \\
        InternVL2 & 1B/0.5B & 2.77 & 0.17 & 12.68 & \hgreen{29.23} \\
        InternVL2 & 1B/1.5B & 2.36 & 0.15 & 10.80 & \hgreen{30.03} \\
        \midrule
        LongVA & Uniform & 128 & 1 & 465.44 & 24.41 \\
        LongVA & 1B/0.5B & 2.77 & 0.02 & 10.07 & \myred{23.18} \\
        LongVA & 1B/1.5B & 2.36 & 0.02 & 8.58 & \myred{23.85} \\
        \midrule
        LLaVA-N-i & Uniform & 16 & 1 & 62.78 & 24.37 \\
        LLaVA-N-i & 1B/0.5B & 2.77 & 0.17 & 10.86 & \myred{24.20} \\
        LLaVA-N-i & 1B/1.5B & 2.36 & 0.15 & 9.26 & \myred{24.26} \\
        \Xhline{1.0pt}
        \end{tabular}
    \vspace{-0.1cm}
    \caption{
    \textbf{Evaluations results with selector adoption.}
   }
    \label{tab:selector_bench}
    \vspace{-0.1cm}
\end{table}
We further apply the selector to other LVLMs to explore whether the extracted core frames can be effectively generalized to other models in a zero-shot manner. As shown in Tab.~\ref{tab:selector_bench}, we evaluated the method's performance on GPT-4o and several open-source LVLMs. The results demonstrate performance improvements and a reduction in the number of input frames, with the frame input reduced by approximately $15\%$ on both GPT-4o and InternVL2. For the other two models, experiments show that introducing the selector leads to a slight loss in performance, but substantial gains in frame input. Notably, LongVA achieved a $98\%$ reduction in frame input, which highlights that our proposed selector still aids in reducing computational overhead for reasoning in LLMs, as a plug-and-play module.
\section{Conclusion}
\label{sec:concl}
In this paper, we presented \textit{VideoEspresso}, a novel dataset designed to enhance video reasoning by addressing the limitations of existing datasets in terms of scale and granularity. Our approach employs semantic-aware key-frame extraction and leverages GPT-4o to generate fine-grained VideoQA pairs with Chain-of-Thought evidence. By integrating a Hybrid LVLMs Collaboration framework, we achieve cost-effective and accurate video reasoning, outperforming baseline models on the majority of tasks across our proposed benchmark. \textit{VideoEspresso} sets a new starting point in video reasoning, offering rich annotations that facilitate advanced multimodal understanding. We hope our contributions can facilitate future exploration and development of more sophisticated models capable of tackling complex video reasoning challenges.
{
    \small
    \bibliographystyle{ieeenat_fullname}

}
\clearpage
\appendix
\section{Details of VideoEspresso}
\label{sec:supp_detail_data}
\begin{table}[h]
    \centering
    \resizebox{0.45\textwidth}{!}{%
    \begin{tabular}{lccc}
    \toprule
    \textbf{Benchmark} & \textbf{Core Frames} & \textbf{CoT} & \textbf{\# Questions} \\
    \midrule
    How2QA~\cite{li2021value} & \myred{\ding{55}} & \myred{\ding{55}} & 2,852 \\
    ActivityNet-QA~\cite{yu2019activitynet} & \myred{\ding{55}} & \myred{\ding{55}} & 8,000 \\
    NExT-QA \cite{xiao2021next} & \myred{\ding{55}} & \myred{\ding{55}} & 8,564 \\
    MovieChat~\cite{song2023moviechat} & \myred{\ding{55}} & \myred{\ding{55}} & 13,000\\
    TVQA \cite{lei2018tvqa} & \myred{\ding{55}} & \myred{\ding{55}} & 15,253 \\
    MSRVTT-QA~\cite{xu2017video} & \myred{\ding{55}} & \myred{\ding{55}} & 72,821 \\
    VideoCoT~\cite{wang2024videocot} & \myred{\ding{55}} & \hgreen{\textbf{T}} & 11,182 \\
    \midrule
    \textbf{\textit{VideoEspreeso}} & \hgreen{\ding{51}} & \hgreen{\textbf{T\&V}} & \textbf{203,546} \\
    \bottomrule
    \end{tabular}
    }
    \caption{\textbf{Dataset comparison} between videoQA datasets. \hgreen{\textbf{T}} and \hgreen{\textbf{V}} represent the textual and visual elements in the CoT, respectively.
    }
    \label{table:comparison}
\end{table}
\textbf{Dataset Comparison.} Existing VideoQA datasets~\cite{li2021value,yu2019activitynet,xiao2021next,song2023moviechat,lei2018tvqa,xu2017video,wang2024videocot} are limited by manual annotations, making it challenging to scale up to meet the demands of LVLM training. In contrast, our proposed dataset, \textit{VideoEspresso}, contains over 200K question-answer pairs (Tab.~\ref{table:comparison}), significantly enhancing the dataset scale. Moreover, we annotate highly relevant core frames within the videos, providing a fine-grained representation of temporal information. While VideoCoT~\cite{wang2024videocot} only introduces text-level chains of thought (CoT), we address the gap in previous work by incorporating visual elements into CoT process.

\begin{figure*}[t]
   \centering
   \includegraphics[width=\textwidth]{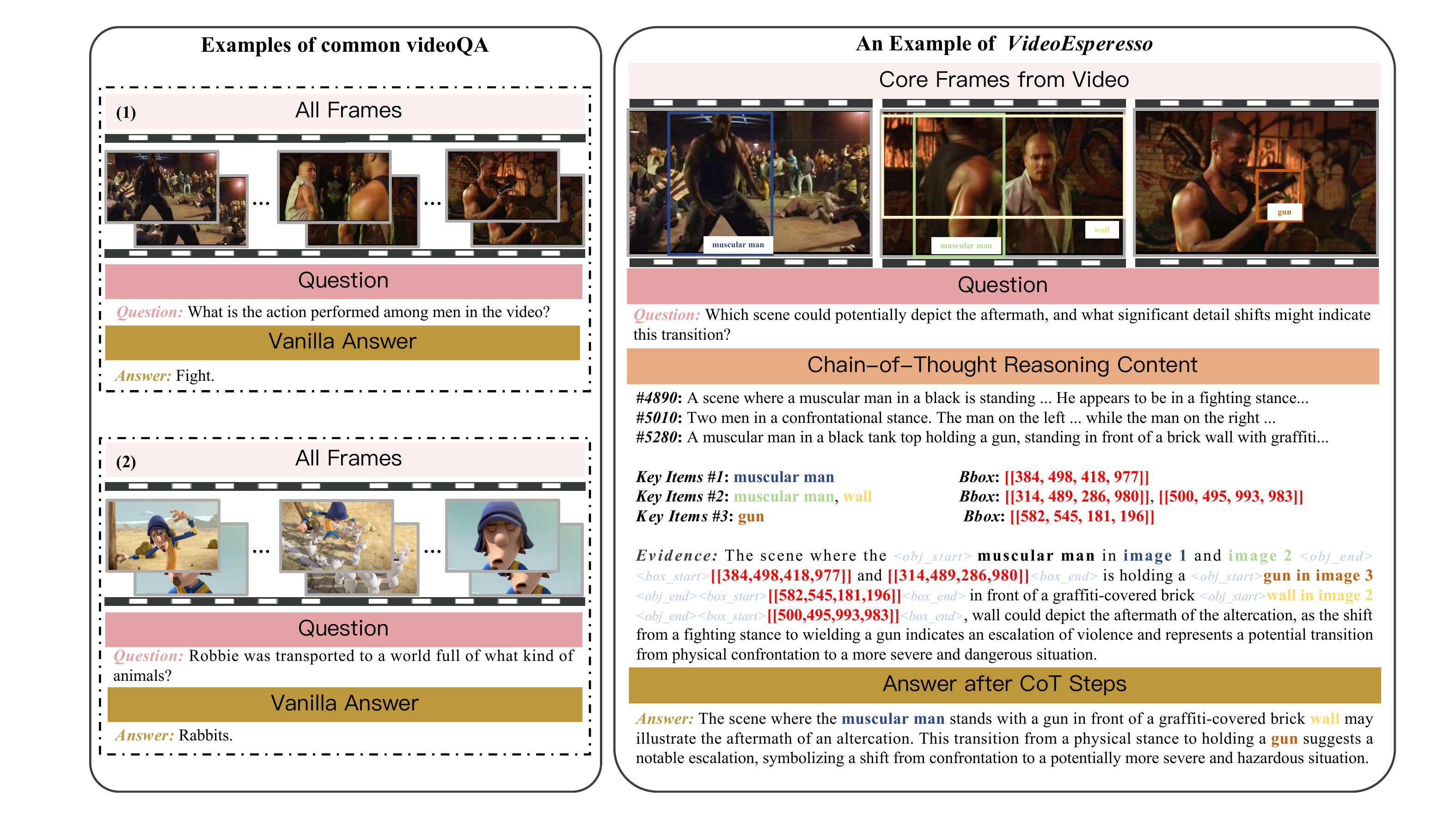}
   \caption{\textbf{Comparison} between \textit{VideoEspresso} and other VideoQA dataset.}
   \label{fig:example}
\end{figure*}

\begin{table}[t]
    \centering
    \resizebox{0.45\textwidth}{!}{%
    \begin{tabular}{lcc}
    \toprule
     \textbf{Task} & \textbf{\# Train Set} & \textbf{\# Test Set} \\
    \midrule
Causal Inference & 87,009 & 426 \\
Contextual Interpretation & 20,057 & 109 \\
Event Process & 29,227 & 174 \\
Interaction Dynamics & 7,322 & 62 \\
Behavior Profiling & 660 & 57 \\
Emotional Recognition & 3,505 & 65 \\
Influence Tracing & 5,749 & 72 \\
Role Identification & 9,134 & 63 \\
Narrative Structuring & 3,940 & 62 \\
Thematic Insight & 10,650 & 61 \\
Situational Awareness & 1,018 & 50 \\
Cooking Steps & 276 & 53 \\
Ingredient Details & 22,552 & 98 \\
Traffic Analysis & 1,065 & 30 \\
\midrule
Total & 202,164 & 1,382 \\
\bottomrule
    \end{tabular}
    }
    \caption{\textbf{Tasks distribution} and \textbf{dataset split} in \textit{VideoEspresso}.
    }
    \label{table:test_samp}
\end{table}
\textbf{Dataset details.} As shown in Tab.~\ref{table:test_samp}, \textit{VideoEspresso} comprises 14 tasks, with the training and testing sets divided according to specific proportions. The detailed question design for each task is presented in Tab.~\ref{tab:tasks}.
As shown in Fig.~\ref{fig:example}, traditional videoQA datasets sample all frames of a video at equal intervals. In contrast, \textit{VideoEspresso} only focuses on the core frames of the video, which are highly relevant to the question. Unlike conventional videoQA tasks, which predominantly focus on querying actions or participants within the video, our dataset prioritizes the fine-grained logical reasoning, requiring a deeper understanding of complex temporal and contextual relationships. Moreover, the analysis of multimodal evidence integrated within the Chain-of-Thought reasoning process enhances both the accuracy and robustness of the generated answers, ensuring they are substantiated by comprehensive contextual understanding.

\begin{table}[ht]
    \centering
    \setlength\tabcolsep{10pt}
    \resizebox{1.0\linewidth}{!}{
        \begin{tabular}{l|cc}
        \textbf{config} & \textbf{Stage1} & \textbf{Stage2} \\
        \Xhline{1.0pt}
        input resolution & 224 & 224 \\
        max token length & 6144 & 6144 \\
        LoRA & \multicolumn{2}{c}{True} \\
        weight ratio & \multicolumn{2}{c}{0.02} \\
        learning rate schedule & \multicolumn{2}{c}{cosine decay} \\
        learning rate & 2e-5 & 1e-5 \\
        batch size & \multicolumn{2}{c}{16} \\
        warmup epochs & 0.03 & 0.03 \\
        total epochs & 1 & 1 \\
        \end{tabular}
    }
    \vspace{-0.3cm}
    \caption{
        \textbf{Training Hyperparameters} for different stages.
    }
    \label{tab:hyperparameters} 
\end{table}

\begin{table*}[!t]
    \centering
    \renewcommand{\arraystretch}{1.2}
    \setlength\tabcolsep{1pt}
    \resizebox{1\textwidth}{!}{
        \begin{tabular}{l c}
       \rowcolor{summary_color} \multicolumn{2}{c}{\textbf{Logical Reasoning}} \\
\midrule
\textbf{Causal Inference} & \textit{How did the actions of the robot and display on the screen contribute to the successful resolution in the control room?} \\
\midrule
\textbf{Contextual Interpretation} & 
\textit{How does the presence of the small cat and George's exploration relate to the chef's activities?} \\
\midrule
\textbf{Event Process} & \textit{What transition do the rabbits experience from the time the moon rose to when they drift off to sleep?} \\
\addlinespace
\rowcolor{perception_color} \multicolumn{2}{c}{\textbf{Social Understanding}} \\
\midrule 
{\textbf{Interaction Dynamics}} & \textit{Considering the atmosphere and expressions depicted, what can be concluded about the progression of the interaction between the man and the woman?} \\
\midrule
{\textbf{Behavior Profiling}} & \textit{Discuss how the actions of the baby triceratops with different dinosaurs reveal aspects of its behavior and the responses of the other dinosaurs.} \\
\midrule
\textbf{Emotional Recognition} & \textit{How does the emotional journey of the small purple dinosaur from feeling lost to excitement tie into the group's decision to explore the cave?} \\
\midrule
\textbf{Influence Tracing} & \textit{How did the presence of the dolphin and the sea monster influence the dinosaurs' experience at the waterbody?} \\
\addlinespace
\rowcolor{reasoning_color} \multicolumn{2}{c}{\textbf{Discourse Comprehension}} \\
\midrule
\textbf{Role Identification} & \textit{How does the woman's role in coordinating town safety relate to the device's activation with a green checkmark and an orange flame?} \\
\midrule
\textbf{Narrative Structuring} & \textit{Considering the changes between the two frames, what can you infer about the narrative progression between the two depicted scenes?} \\
\midrule
\textbf{Thematic Insight} & \textit{How do the  changing production logos contribute to the thematic preparation for the viewer before the main storyline begins?} \\
\midrule
\textbf{Situational Awareness} & \textit{Based on the sequence of events, how does the situation described contribute to the visual effect observed in the third frame?} \\
\addlinespace
\rowcolor{navigation_color} \multicolumn{2}{c}{\textbf{Reality Application}} \\
\midrule
\textbf{Cooking Steps} & \textit{Considering the sequence of actions, what cooking technique is being employed, and how is it crucial for the fried chicken?} \\
\midrule
\textbf{Ingredient Details} & \textit{If the person is preparing chili con carne, what is the purpose of the liquid being poured into the pan?} \\
\midrule
\textbf{Traffic Analysis} & \textit{Analyze the potential destinations of the visible vehicles based on their types and cargo as inferred from the images.} \\
\bottomrule
        \end{tabular}
    }
    \vspace{-0.2cm}
    \caption{
    \textbf{Our proposed task categories with question prototypes.}
   }
    \label{tab:tasks}
    \vspace{-0.3cm}
\end{table*}

\section{Training Implementation}
\label{sec:supp_implement}
The hyperparameters used at different training stages are listed in Tab.~\ref{tab:hyperparameters}, following LLaVA-Next architecture~\cite{liu2023llava,li2024llava}. During both stage, we leverage diverse instruction data and integrate LoRA modules~\cite{hu2021lora} into the LLM with a rank of 16, an alpha value of 32, and a dropout rate of 0.1. Flash attention~\cite{dao2022flashattention} is applied to accelerate the training process.

\section{Prompt Details}
\label{sec:supp_prompt}
\begin{figure}[t]
   \centering
   \includegraphics[width=0.45\textwidth]{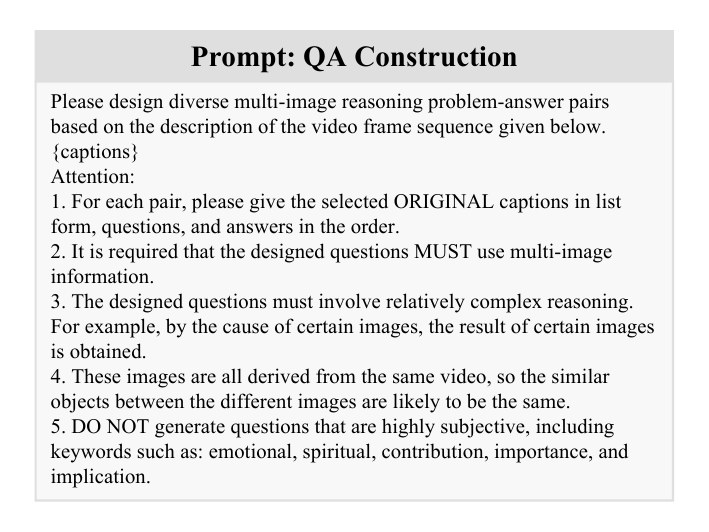}
   \caption{\textbf{QA-Construction Prompt.}}
   \label{fig:prom_1}
\end{figure}

\begin{figure}[t]
   \centering
   \includegraphics[width=0.45\textwidth]{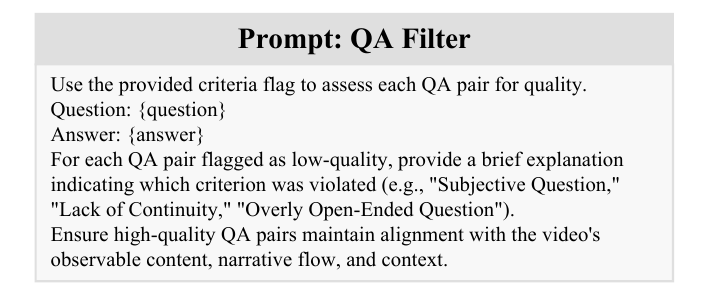}
   \caption{\textbf{QA-Filter Prompt.}}
   \label{fig:prom_2}
\end{figure}

\begin{figure}[t]
   \centering
   \includegraphics[width=0.45\textwidth]{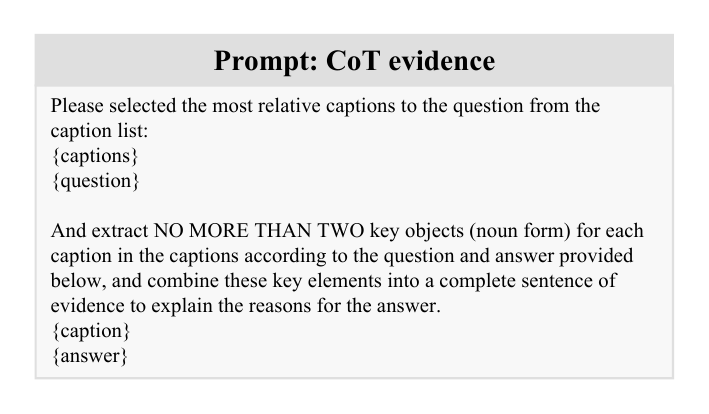}
   \caption{\textbf{CoT-Evidence Construction Prompt.}}
   \label{fig:prom_3}
\end{figure}

\begin{figure}[t]
   \centering
   \includegraphics[width=0.47\textwidth]{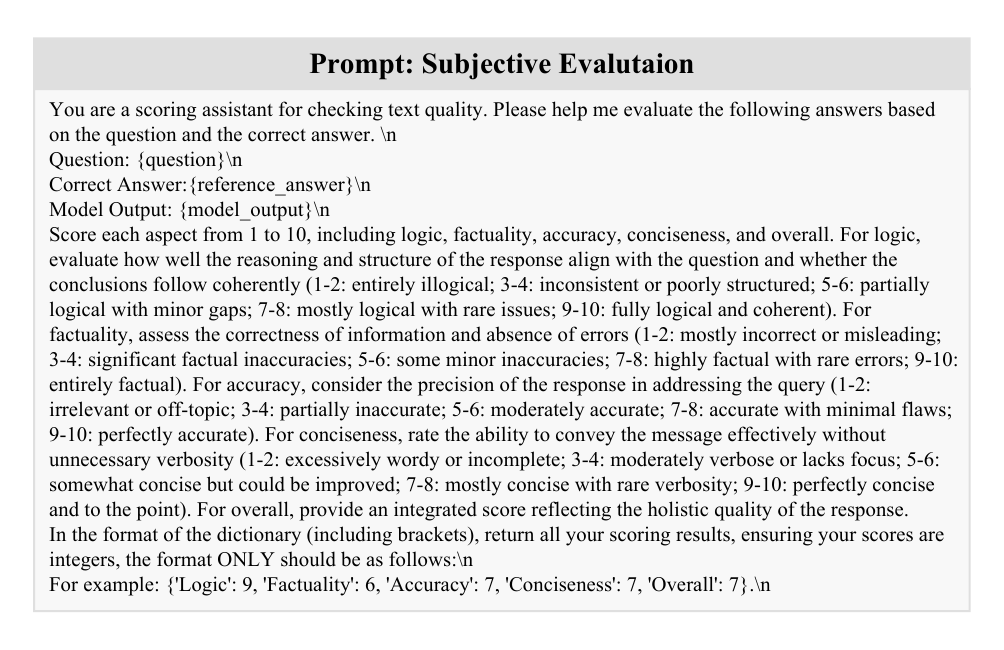}
   \caption{\textbf{Subjective Evaluation Prompt.}}
   \label{fig:prompt_subeval}
\end{figure}

In this section, we present the complete set of prompts utilized in the data generation pipeline, alongside those employed for subjective evaluation. Specifically, these include the prompt designed for QA construction in Fig.~\ref{fig:prom_1}, the prompt aimed at filtering low-quality QA pairs in Fig.~\ref{fig:prom_2}, the prompt used for constructing CoT evidence in Fig.~\ref{fig:prom_3}, and the prompt applied for subjective evaluation in Fig.~\ref{fig:prompt_subeval}.

\section{Evaluation Analysis}
\label{sec:supp_eval}
\textbf{Construction of Test set.} For all questions, we devised three distractor options that maintain consistent contextual relevance and similar linguistic structures to the correct answer while presenting distinct factual inaccuracies, thereby enhancing the robustness of the objective process. Furthermore, to mitigate potential biases arising from significant token-length disparities in the second step of the objective evaluation, we employed GPT-4o~\cite{openai2024gpt4o} to standardize the length and ensure a balanced distribution across all answer options for each question (shown in Fig.~\ref{fig:ope_len}).

\textbf{Details of Objective Evaluation.} As illustrated in Algorithm~\ref{alg:obj}, our objective evaluation is divided into two distinct steps. In the first step, the semantic similarity between the model's output $O$ and the reference answer $R$ is computed. If the similarity score $S_R$ falls below the predetermined threshold $tau = 80\%$, the output is deemed incorrect. In the second step, a set of three carefully selected confounding distractors $\{D_1, D_2, D_3\}$ is introduced for each reference answer. The semantic similarity $S_{D_i}$ between the model's output and each distractor is then computed. If any distractor's similarity score $S_{D_i}$ exceeds $S_R$, the output is categorized as incorrect. Only outputs meeting the criteria in both steps are ultimately classified as correct.

\begin{figure}[t]
   \centering
   \includegraphics[width=0.45\textwidth]{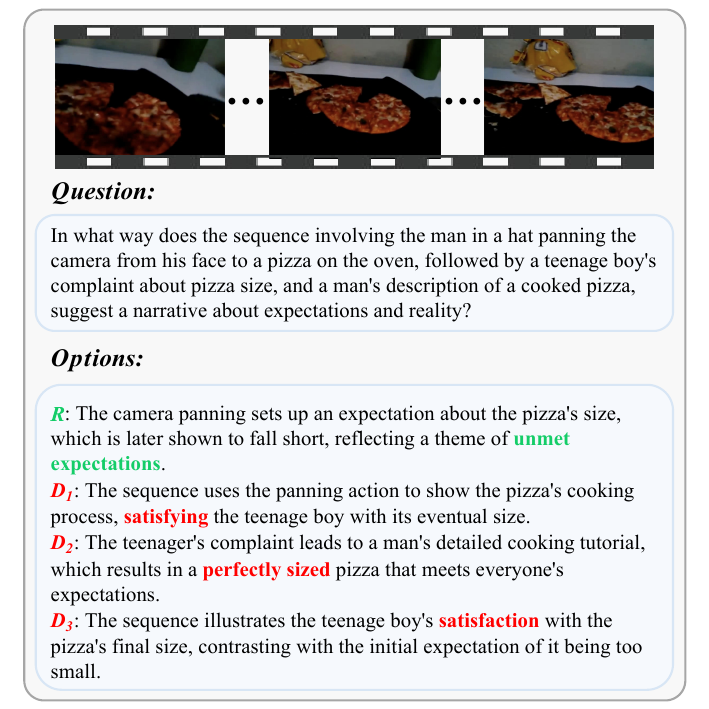}
   \caption{\textbf{Example of test set.} \hgreen{\textit{\textbf{$R$}}} represent the Reference Answer, while \myred{\textit{\textbf{$D_{i}$}}} stand for the $i$-th Distractor.}
   \label{fig:example_opt}
\end{figure}

\begin{figure}[t]
   \centering
   \includegraphics[width=0.45\textwidth]{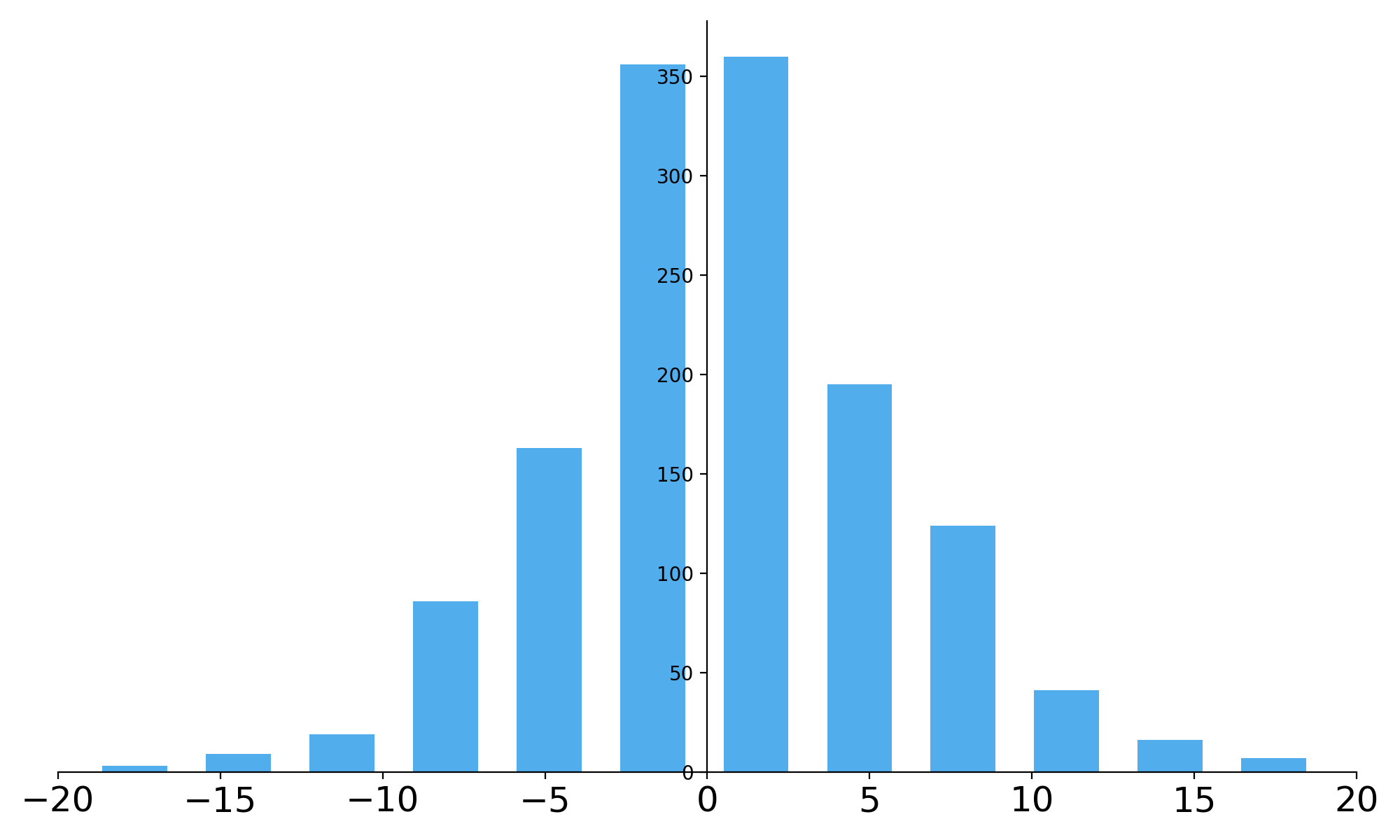}
   \caption{\textbf{The Distribution} of token length disparities between reference answers and the longest distractor option.}
   \label{fig:ope_len}
\end{figure}

\textbf{Analysis of test set.} 
As depicted in Fig.~\ref{fig:example_opt}, we present the example of reference answers and distractor options within the test set. The figure highlights factual inaccuracies in the distractor options using red annotations, while the correct answers are distinctly marked in green for clarity and emphasis.
The token length disparities between reference answers and the longest distractor option, as shown in Fig.~\ref{fig:ope_len}, predominantly are confined to the interval $[-10, +10]$, indicating that the disparity in length between correct answers lie within is relatively minor. The distribution shows near symmetry along the y-axis, indicating a balanced pattern: in about half of the cases, reference answers are longer than distractors, while in the remaining cases, distractors are longer.

\begin{algorithm}[t]
\caption{Objective Evaluation for Open-Ended Output}
\label{alg:obj}
\begin{algorithmic}[1]
\Require Model output $O$, reference answer $R$, 
threshold $\tau = 80\%$, distractors $\{D_1, D_2, D_3\}$
\Ensure Evaluation result: Correct or Incorrect

\LeftComment{Step 1: Semantic Similarity Assessment}
\State Compute semantic similarity $S_R = \text{Sim}(O, R)$
\If{$S_R < \tau$}
    \State \textbf{Return: Incorrect}
\EndIf

\LeftComment{Step 2: Confounding Distractor Analysis}
\For{each distractor $D_i$ in $\{D_1, D_2, D_3\}$}
    \State Compute semantic similarity $S_{D_i} = \text{Sim}(O, D_i)$
    \If{$S_{D_i} > S_R$}
        \State \textbf{Return: Incorrect}
    \EndIf
\EndFor

\State \textbf{Return: Correct}
\end{algorithmic}
\end{algorithm}

\section{Case Study}
\label{sec:supp_casestudy}
\begin{figure}[t]
   \centering
   \includegraphics[width=0.45\textwidth]{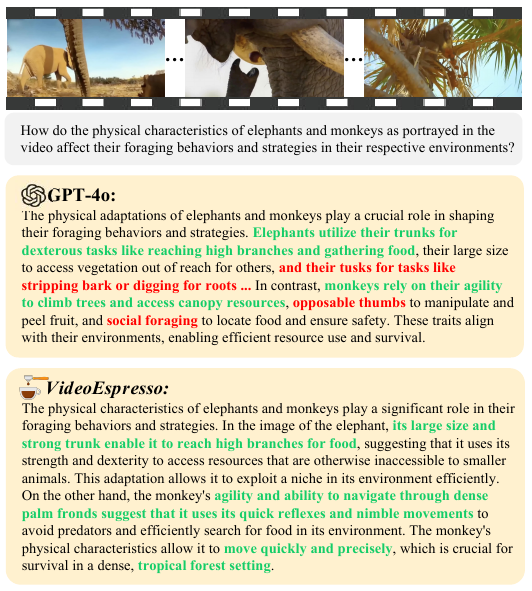}
   \caption{\textbf{Example of over-analysis with GPT-4o.}}
   \label{fig:case1}
\end{figure}

\begin{figure}[t]
   \centering
   \includegraphics[width=0.45\textwidth]{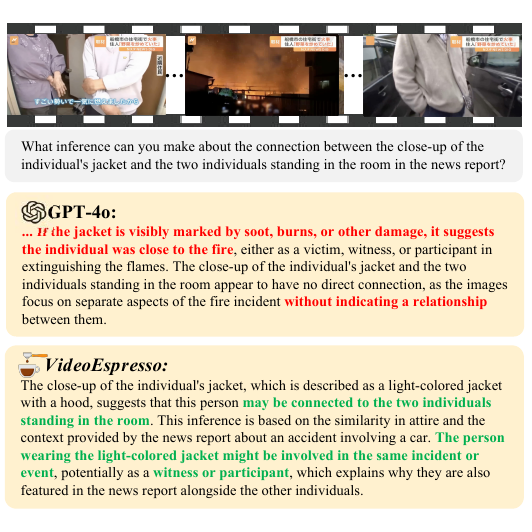}
   \caption{\textbf{Example of Non-factual response with GPT-4o.}}
   \label{fig:case2}
\end{figure}

Leveraging \textit{VideoEspresso}, the LVLM demonstrates superior performance in fine-grained reasoning tasks for video understanding. As shown in Fig.~\ref{fig:case1}, while GPT-4o generates seemingly rich and plausible answers, including detailed analyses of elephant and monkey behaviors, it incorporates a significant amount of video-irrelevant information, such as ``\textit{and their tusks for tasks like stripping bark or digging for roots}" and ``\textit{social foraging}". This diminishes the proportion of visually grounded outputs. This issue is more pronounced in Fig.~\ref{fig:case2}, where GPT-4o performs extensive analysis based on erroneous reasoning due to its failure to account for fine-grained inter-frame relationships, e.g., ``\textit{If the jacket is visibly marked by soot, burns, or other damage, it suggests the individual was close to the fire.}" From these examples, we underscore the critical role of visual information in video reasoning QA tasks.

\end{document}